\documentclass[10pt,twocolumn,letterpaper]{article}

\usepackage{cvpr}
\usepackage{times}
\usepackage{epsfig}
\usepackage{amsmath}
\usepackage{amssymb}
\usepackage{xcolor}
\usepackage{enumitem}
\usepackage{hyperref}

\newcommand{\UTD}{University of Texas at Dallas}

\begin{document}

\title{Human-Machine Comparison for Cross-Race Face Verification: Race Bias at the Upper Limits of Performance?}

\author{G\'{e}raldine~Jeckeln\\
The University of Texas at Dallas\and  Selin Yavuzcan\\
The University of Texas at Dallas\and   Kate A. Marquis\\
The University of Texas at Dallas \and Prajay Sandipkumar Mehta\\
The University of Texas at Dallas \and Amy N. Yates\\
National Institute of Standards and Technology \and P. Jonathon Phillips\\
National Institute of Standards and Technology \and Alice J. O’Toole\\
The University of Texas at Dallas}


\maketitle
\thispagestyle{empty}

\begin{abstract}
Face recognition algorithms perform more accurately than humans in some cases---though humans and machines both show race-based accuracy differences. As algorithms continue to improve, it is important to continually assess their race bias relative to humans. We constructed a challenging test of ``cross-race'' face verification and used it to compare humans and two state-of-the-art face recognition  systems.  Pairs of same- and different-identity faces of White and Black individuals were selected to be difficult for humans and an 
open-source implementation of the ArcFace face recognition algorithm from 2019 \cite{deng2019arcface}. Human participants (54 Black; 51 White) judged whether face
pairs showed the same identity or different identities on a 7-point Likert-type scale. Two top-performing face recognition  systems from the Face Recognition Vendor Test-ongoing \cite{FRVTongoing} performed the same test. By design, the test proved challenging for humans as a group, who performed above chance, but far less than perfect. Both state-of-the-art face recognition  systems scored perfectly (no errors), consequently with equal accuracy for both races. We conclude that state-of-the-art systems for identity verification between two frontal face images of Black and White individuals can surpass the general population. Whether this result generalizes to challenging in-the-wild images is a pressing concern for deploying face recognition systems in unconstrained environments.
\end{abstract}

\section{Introduction}

Machine-based face recognition is commonly bench-marked against human performance \cite{phillips2014comparison}. Prior to the advent of deep learning in 2012  \cite{krizhevsky2012imagenet}, human performance was considered the gold standard for face recognition. By 2007 \cite{o2007face}, state-of-the-art face recognition algorithms were more accurate than humans for constrained frontal images.
As face recognition algorithms have continued to improve in the era of deep networks, human-machine comparisons have evolved to include more sophisticated
human participants (e.g., trained professional forensic facial examiners, super-recognizers, 
forensically-trained fingerprint examiners) and more challenging facial comparisons \cite{phillips2018face}. 
In a 2018 test of frontal face-image pairs, curated to be challenging for untrained individuals and for
previous algorithms \cite{phillips2018face},
the best algorithms  performed substantially more accurately than untrained people. These algorithms also proved closely competitive with highly trained forensic facial examiners.

Despite the upward trend in performance for algorithms, multiple studies show that algorithms 
perform at different levels of accuracy for faces of different races 
(cf., \cite{cavazos2020accuracy,grother2019face}). This raises obvious concerns about the fairness of deploying face recognition systems in application scenarios.
Race bias in the performance of face recognition systems has been reported for decades and spans widely different types of algorithms (see \cite{cavazos2020accuracy} for a review). 
Recent studies indicate that this bias persists in the deep learning era \cite{cavazos2020accuracy,grother2019face,S_2019_CVPR_Workshops,krishnapriya2020issues,cook2019demographic,howard2019effect,Howard2021ReliabilityAV}.
For example, an ongoing international test of face recognition algorithms at the National Institute of Standards and Technology (NIST) \cite{grother2019face},
found that nearly all algorithms tested ($n=189$) showed performance differences as a function of the race of the face. 
Notwithstanding ubiquitous findings of bias, the direction of bias (e.g., which races were most/least accurately identified) was not consistent across algorithms. This  complicates decisions about whether/how to use face recognition algorithms in multi-race settings.

Similar to algorithms, people recognize 
faces of different races with different levels of accuracy.
The ``other-race effect'' (ORE) for face recognition refers to long-standing findings indicating that people are more accurate at recognizing faces of their own race than faces of other races (cf. \cite{malpass1969recognition,meissner2001thirty}). This raises questions about the reliability and equity of human face recognition in multi-race settings.

The combined factors of human and machine bias take on added importance now, as machines and humans both contribute regularly to face identification decisions in consequential applied scenarios (e.g., law enforcement, security). Despite decades of human-machine face recognition comparison (e.g., \cite{Kumar_attributes,o2007face,phillips2014comparison,phillips2018face}), there is only one human-machine comparison that examined  cross-race face verification \cite{phillips2011other}. This study analyzed systems submitted to the FRVT 2006 \cite{Phillips:2009zl}---prior to the development of deep learning algorithms. To test for an ``algorithm other-race effect,'' the authors created a Western fusion algorithm (a fusion of algorithms developed in Western countries: France, Germany, and the United States) and  an East Asian fusion algorithm (a fusion of algorithms developed in East Asian countries: China, Japan, and Korea). The Western fusion algorithm performed more accurately on Caucasian faces\footnote{Terms used to describe racial/ethnic groups are used as they appear in the original published papers.} than East Asian faces, whereas the East Asian fusion algorithm performed more accurately on East Asian faces than on Caucasian faces. In a direct human-to-algorithm comparison on a curated subset of faces, humans showed a pattern of results that mirrored a classic ORE--with a small advantage for ``own-race'' faces. Notably, both the Western fusion algorithm and the East Asian fusion algorithm showed precipitous performance drops for recognition of ``other-race'' faces. Thus, the performance of both humans and algorithms was race-biased, but human performance was more stable than algorithm performance. Also, of note for the present work, algorithms performed less accurately than humans in every condition.

The goal of the present study was to compare state-of-the-art face recognition systems to humans on an identity verification test for the faces of Black and White individuals. We created a challenging cross-race face identification test using an open-source deep convolutional neural network (DCNN) and a human experiment to pre-screen face pairs. Next, we administered this test to untrained participants and further refined the stimulus set to include items that people found especially challenging. Face pairs were further sub-sampled to equate human performance across participant and stimulus races.  
The resulting test was administered to two  face recognition systems for a human-machine comparison. 

Designing and testing our cross-race test is the first step to measuring the cross-race abilities of forensic facial examiners when they have access to their laboratory, tools, and methods. The forensic community refers to this as a black-box test. Such a study would build on previous work  that measured the abilities of facial examiners, super-recognizers, and algorithms \cite{phillips2018face}. A perceptual test supports conducting a cross-race black-box study~\cite{Yates2023}.

The contributions of this study are:
\begin{itemize}[nosep]
     \item First  cross-race face recognition comparison between humans and machines in the deep learning era. 
     \item Demonstration that two state-of-the-art algorithms performed perfectly on the faces of the Black and White individuals  on a test that showed poor performance for untrained human participants of both races. 
    \item Verification that the untrained humans we tested were typical of untrained humans tested in past human-machine comparisons \cite{phillips2018face}. This further validates the finding of machine superiority for cross-race face identification. 
    \item Development of a challenging cross-race test for the faces of Black and White individuals. The  difficulty of this test makes it suitable for evaluating cross-race face identification with experts (e.g., professional forensic facial examiners, and super-recognizers).  
 \end{itemize}

This paper is organized as follows: First, we describe the test construction process, and the first step of stimulus 
pre-screening with an open-source DCNN (Section \ref{test_construction}). Second, we describe the experiments (Section \ref{sec:experiments}). These include: 1.) a cross-race test administered to humans, which is used to further pre-screen face pairs and provide human performance for comparison with face recognition systems; 2.) a cross-race test administered to two state-of-the-art face recognition systems;  
and 3.) a previously normalized face recognition test administered to humans verify the comparability of participants in this experiment relative to a human-machine comparison study \cite{phillips2018face}. Third,
we report all human and machine results 
(See Section \ref{sec:results}). Fourth, we discuss the results and their implications for deploying face recognition systems in diverse environments.

\section{Constructing a Cross-Race Test}\label{test_construction}

\textbf{Face images.}
To create the cross-race identity verification test, we searched the Notre Dame Million Biometric Sample data set \cite{Phillips:2016aa} for frontal face images of Black and White individuals. The search produced 3102 images from 117 Black individuals and 122,728 images from 2411 White individuals. The ratio of Black to White individuals and images is representative of the subjects that volunteered to  participate in the Notre Dame Million Biometric Sample data collection. Next, we assembled pairs of same- and different-identity images for each race. Due to the limited number of face images of Black individuals, we compiled those pairs first and then assembled pairs for the White individuals by matching demographic (age and gender) and image variables (environment) to the image pairs of the Black individuals. Race and gender information was obtained through the self-identification of the subjects at the time the images were taken.

\begin{figure}
    \centering
    \begin{tabular}{c}
    Cross-Race Test: Image Screening and Testing Process\\ 
            \includegraphics[width=0.50\textwidth]{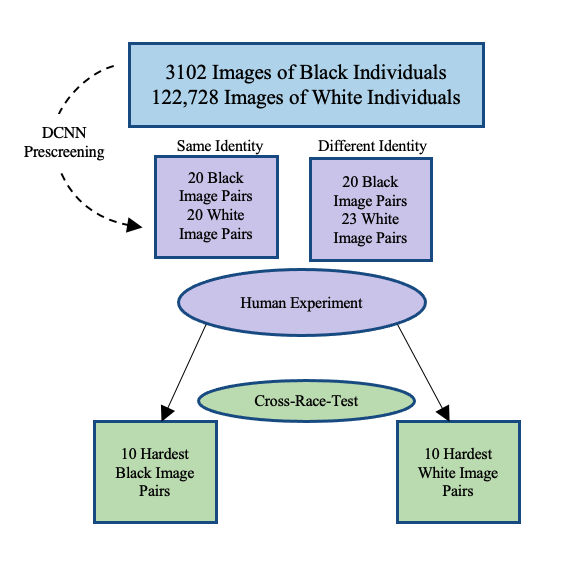} 
    \end{tabular}
    \caption{\small Flow chart overview of the sampling of the database images. 
    The database contained 3102 Black and 122,728 White identities (blue box). A first  DCNN pre-screening was done based on the similarity of DCNN feature codes of face pairs (see main text), reducing this large number of identities to 83 Black and White identity pairs (purple boxes). These face pairs were presented in the cross-race human experiment to identify the most challenging identity pairs. A subset of untrained human participants was used to determine a highly challenging set of (10 Black and 10 White) identity pairs (green boxes).}
    \label{fig:flow_chart}
\end{figure}

\textbf{Overview of stimulus pre-screening.}
The challenge was to find pairs of face images of Black and White individuals that would be difficult for humans and machines to identify. Therefore, we pre-screened face-image pairs in multiple steps. An overview of the pre-screening procedure appears in Figure \ref{fig:flow_chart}. 
To summarize, we first employed an open-source DCNN \cite{deng2019arcface} to find difficult image pairs. These were defined as different-identity image pairs with similar DCNN feature codes and same-identity image pairs with dissimilar DCNN feature codes. 
DCNN prescreening  reduced the number of images from the available database to the 83 pairs utilized in the 
human cross-race experiment. Next, we tested untrained Black and White participants on an identity verification task with the pre-screened pairs. This cross-race experiment was used to locate the most difficult face pairs for humans, based on the performance of a subset of participants (26 White and 11 Black participants). These 20 difficult face pairs were retained and used in the human-machine comparisons, which included performance data from all participants on the curated subset of highly challenging face pairs. These difficult items also yielded roughly equal identification accuracy for 
both participant and stimulus race variables, creating a 
set of images with balanced difficulty for face images of
Black and White individuals.

\textbf{Open-source network pre-screening.}\label{network_prescreen}
Face images were processed using the ArcFace network \cite{deng2019arcface}, available through the DeepFace library for Python \cite{serengil2020lightface}. At the time of its release in 2019, the algorithm produced state-of-the-art face verification accuracy \cite{serengil2020lightface}. We chose this network for its level of accuracy among available open-source face recognition algorithms. 

Face-image pairs were constructed separately for different demographic groups (Black/White, female/male). 
{\it Different identity pairs} were created as follows:
For each identity, $A$, an ``identity template'' was generated by averaging the DCNN feature descriptors for each image of that identity. Gender- and age-matched identities were paired to produce identity-template pairs. For each identity-template pair, $A$ and $B$, cosine similarity between the $A$ and $B$ identity templates was computed. The resulting identity-template pairs were ranked from most to least similar. The most similar pairs were retained. Ranked identity-template pairs were inspected visually to ensure that the identities were similar and matched in age group. For each identity-template pair, $A$ and $B$, all possible different-identity image pairs were assembled and cosine similarity between DCNN feature descriptors of the two images was computed. Different identity pairs were ranked from most to least similar, and the most similar pairs were selected. Image pairs were again inspected visually to ensure similar face images. 

{\it Same-identity image pairs} were also selected based on cosine similarity between the feature descriptors for face images, as follows: For each identity, all possible same-identity image pairs were assembled. Next, we computed the cosine similarity for the DCNN features of each face-image pair. The most dissimilar face-image pairs were retained for use in the test. As noted previously, image pairs were inspected visually for quality. 

\begin{figure}[t]
\begin{center}
    Cross-Race Face Matching Task\\ 
   \includegraphics[width=0.8\linewidth]{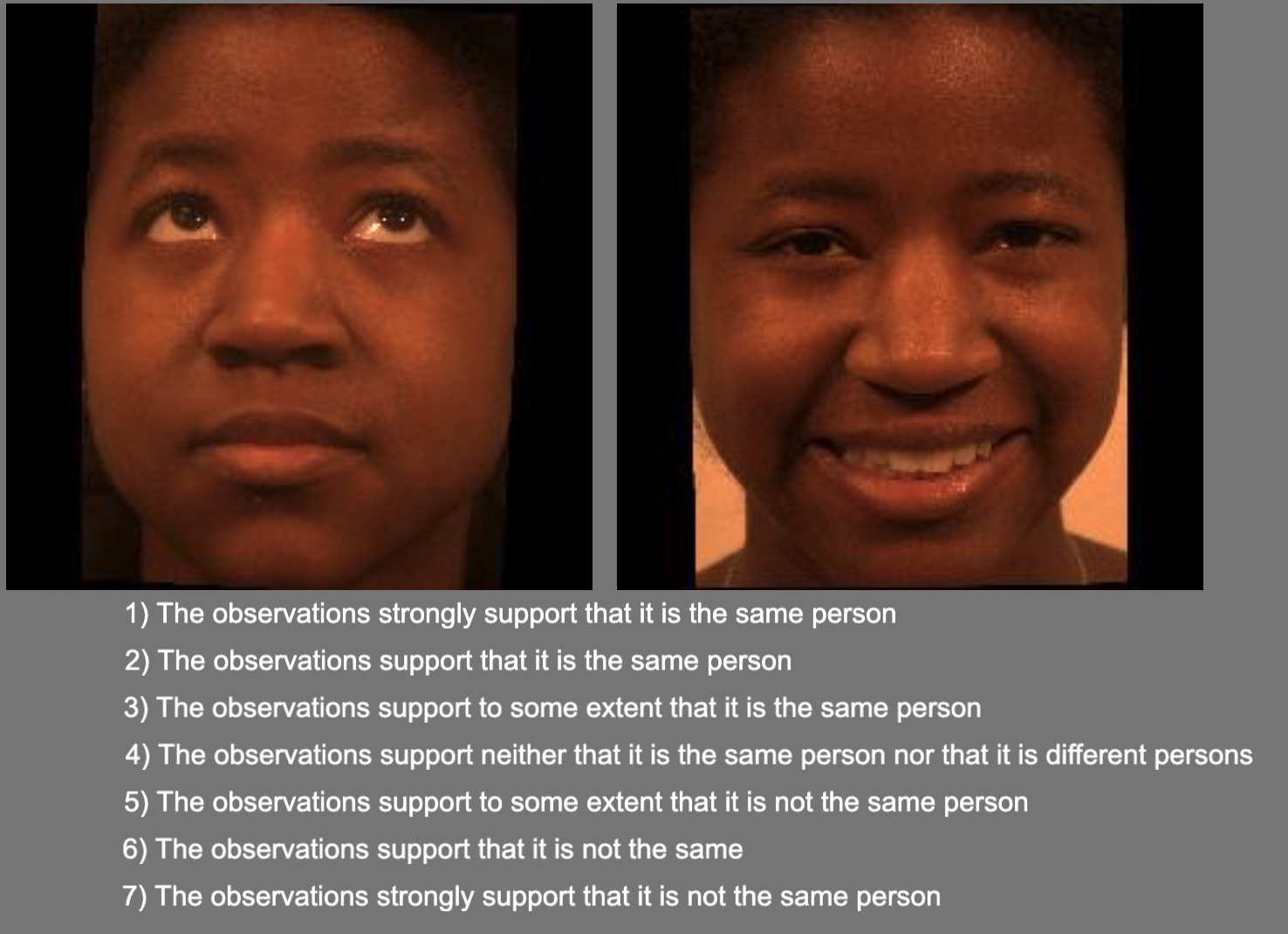}
\end{center}
   \caption{\small Example of an identity pair shown in the Cross-Race Test. Below the pair of faces is a Likert-type scale ranging from 1 (``The observations strongly support that it is the same person'') to 7 (``The observations strongly support that it is not the same person'') from which subjects select if they deem the identities to be the same or different.}
\label{fig:CRE_example_pair}
\label{fig:onecol}
\end{figure}

This procedure yielded 83 image pairs [43 different-identity (21 Black, 22 White) and 40 same-identity (20 Black, 20 White)] (See purple boxes, Figure 1). No identities were repeated in the face-image pairs. An example face pair appears in Figure \ref{fig:CRE_example_pair}.

 Next, we verified that the open-source
 algorithm performed poorly on the
 83 face-image pairs (see Section \ref{test_construction}) selected by computing the area under the Receiver Operating Characteristic (AUC), using the cosine similarity scores for same- and different-identity face image pairs. An AUC of 1.00 represents perfect performance (no errors), whereas an AUC of 0.50 represents random performance.  As expected, the open-source ArcFace network performed at (or near) chance for both conditions (overall AUC = 0.58, White face-image pairs, AUC = 0.50; Black face-image pairs, AUC = 0.65).



\section{Experiments}\label{sec:experiments}

The behavioral part of this study employed two tests: the Cross-Race Test (see Section \ref{test_construction}) 
and a standardized face-matching test called the ``Face Examiner Test'' (FET) \cite{phillips2018face}.   
The two tests were administered, back-to-back, in a single experimental session. Participants completed the Cross-Race Test first and the FET second. The Cross-Race Test served two purposes. First, it provided performance data on the set of face-image pairs pre-selected using the DCNN. Second, it provided human performance data from a subset of  participants ($n=$ 37) to further pre-screen image pairs for the human-machine comparison.\footnote{Participants ($n=$ 105) were tested across two semesters. We used data from the first semester ($n=$ 37) for the human pre-screening of image pairs.}. This human pre-screening  yielded a stimulus set of 20 face-image pairs that were of equal difficulty for both participant races and for both races of face-image pairs--making these face pairs ideal for the human-machine comparison.

\begin{figure}
    \centering
    \begin{tabular}{c}
    \\ 
            \includegraphics[width=0.40\textwidth]{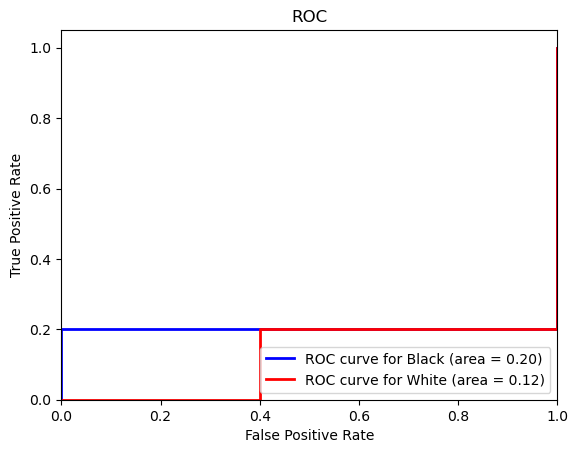} 
    \end{tabular}
    \caption{The receiver operating characteristic (ROC) for ArcFace \cite{deng2019arcface} performing verification on the 10 Black image pairs (blue) and 10 White image pairs (red).
    }
    \label{fig:AUC_20pairs_by_race}
\end{figure}

The FET was used to benchmark participant performance with respect to known population performance statistics. FET performance is known for multiple populations of humans (professional forensic face examiners and reviewers, super-recognizers, fingerprint examiners, and university students). Benchmarks are important in the present case, given that we recruited (untrained) participants from two sources. Machine benchmarks for the FET are also available for four face recognition algorithms (2015-2017, \cite{parkhi2015deep,chen2016unconstrained,ranjan2017all,ranjan2019fast}), making it useful as a reference point for the present human-machine comparisons.
\begin{figure*}
\begin{center}
 \includegraphics[width=1.0\linewidth]{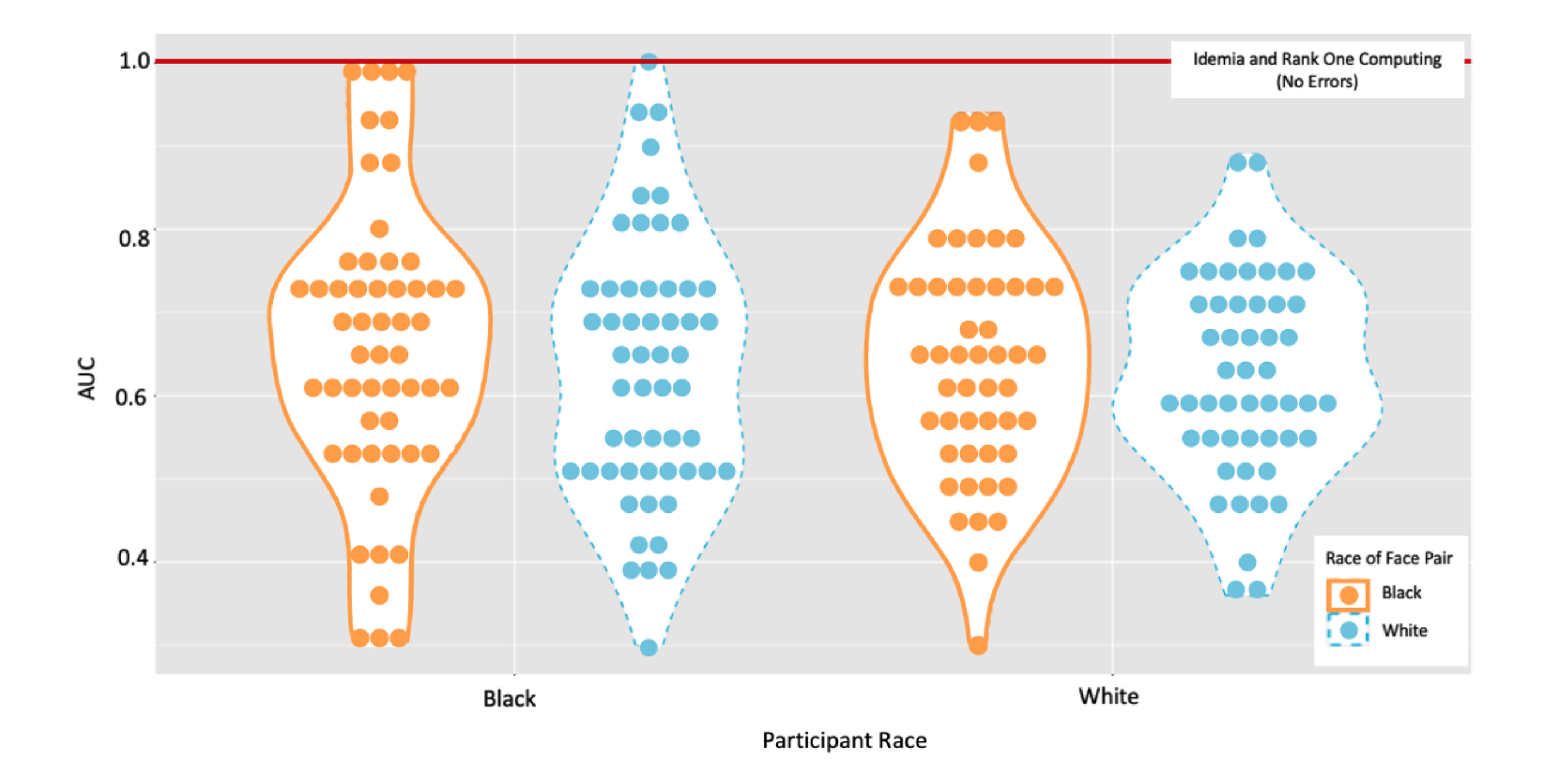}
\end{center}
   \caption{\small These Violin plots show untrained Black and White participant performance on the Cross-Race Test.  
   Each dot represents a participant's AUC score. The $x$-axis depicts the participants' race, and the $y$-axis shows AUC values demonstrating participant performance. 
   The colored outlines indicate the race of face pairs shown in the test, with solid orange for Black and dashed blue for White face pairs. Median AUC values for Black participant performance on the face matching task are as follows: Black face-image pairs = 0.67, White face-image pairs = 0.65. Median AUCs for White participant performance are as follows: Black face-image pairs = 0.64, White face-image pairs = 0.60. The red line indicates the performance of the Idemia and Rank One Computing systems, which both had an AUC  of 1.0.}
\label{fig:short}
\end{figure*}
\subsection{General Methods}

We begin with the standard procedures for the Cross-Race and Face Examiner tests. Because they served different purposes, following the general methods, we describe the experiments separately.

\subsubsection{Participants} 
Volunteers were recruited from 1.) the student population of the \UTD\  using SONA (https://www.sona-systems.com),
and 2.) an online participant recruitment platform, Prolific (www.prolific.co). 
A total of 105 participants (54 Black, 51 White) 
 were tested. Participants self-identified their race and age. 
The \UTD\ Institutional Review Board approved the research. Participants gave informed consent. Student participants were compensated with ``SONA credits'' that count toward the student's coursework at the \UTD.
Payment credits were given to participants recruited
through Prolific. 
Technical difficulties resulted in a small difference in participant numbers for the Cross-Race and 
FET tests with 103 participants completing the FET (49 White and 54 Black participants) and 105 completing the Cross-Race test (51 White and 54 Black participants).
SONA and Prolific participants were combined in all analyses 
based on a statistical comparison between the
two sets of participants that
showed no statistical difference in performance ($p < 0.05$). 

\subsubsection{Procedure}
Participants were tested with an online browser-based interface, programmed in PsychoPy v2022.2.5 \cite{peirce2022building} and administered using Pavlovia (www.pavlovia.org). Tests were completed in one session,  with no time limit for completion.

Face-image pairs were viewed one at a time
in sequence and participants judged whether each pair portrayed the same identity or different identities. Responses were made on a 7-point scale that ranged from ``1: The observations strongly support that it is the same person'' to ``7: The observations strongly support that it is not the same person''. As with the algorithms, accuracy was measured for each participant as AUC (cf., \cite{macmillan2004detection} for AUC application to measure human performance). 

\subsection{Procedures by Test}
\subsubsection{Cross-Race Test: Humans and ArcFace}\label{cross_race_test}
 Participants viewed 83 pairs of face images
(see Section \ref{network_prescreen} and Figure \ref{fig:CRE_example_pair}) and entered their responses. 
Item difficulty was measured using the first group 
of participants tested at the \UTD\ ($n =$ 37; 11 Black, 26 White).\footnote{Note: due to the difficulty of recruiting a sufficient number of Black participants from the \UTD, following this first batch of participants, we began to recruit additional participants using Prolific.}.  The difficulty was assessed using the number of participants who responded incorrectly to the pair (judged ``different'' for same-identity pairs; judged ``same'' for different-identity pairs). Ratings of $5-7$ indicated a ``different-identity'' response; ratings of $1-3$ indicated a ``same-identity response''; a rating of 4 was considered a neutral (don't know) response. 
 The selection process prioritized 
face pairs that produced incorrect responses with high confidence (responses 1-2 and 6-7). The 20 most difficult face-image pairs, equally distributed across race (Black and White identities) and match conditions (same-identity and different-identity pairs), were retained for further analysis.

Returning to the open source algorithm used to pre-screen initially \cite{deng2019arcface}, performance was measured for the 20 image pairs retained. Figure \ref{fig:AUC_20pairs_by_race} shows a receiver operating characteristic (ROC) curve for the performance of ArcFace on the faces of Black and White individuals. In both cases, algorithm responses are systematically incorrect (AUC $<$ 0.50), [Black face image pairs: AUC = 0.20, White face image pairs: AUC = 0.12]. This indicates that the combined human-machine pre-screening process successfully selected face pairs on which a 2019 open-source face recognition algorithm failed.

 For completeness, we tested a second 
open-source face recognition algorithm (Inception-ResNet-v1 network \cite{szegedy2017inception}). This algorithm performed more accurately than ArcFace \cite{deng2019arcface} on these face image pairs, but showed
a strong race bias, with better performance for the 
faces of White individuals (White face image pairs, AUC = 1.00; Black face image pairs, AUC= 0.72).


\subsubsection{Cross-Race Test: Algorithms}

Performance on the Cross-Race Test was evaluated for 
two systems from the FRVT-ongoing evaluation of prototype face recognition systems.
These particular systems 
(Idemia and Rank One Computing submissions) were selected from the top systems under evaluation in 2022-2023 and are  one-to-one face verification systems.

\begin{figure}
    \centering
    \begin{tabular}{c}
        Face Examiner Test \\ \hline
            \includegraphics[width=0.45\textwidth]{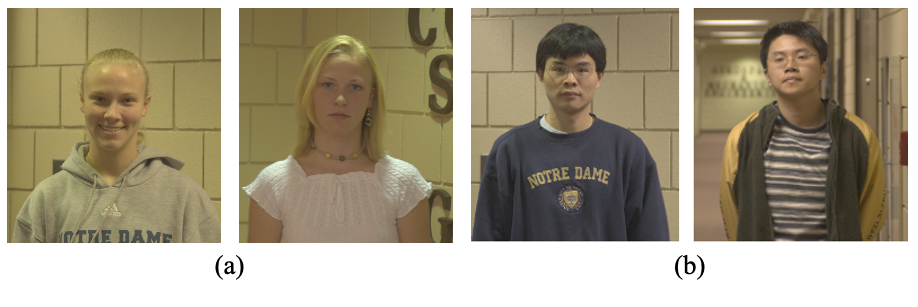} 
    \end{tabular}
    \caption{\small{Example of two image pairs from the Face Examiner Test used to determine if identities presented were a same-identity (left) or different-identity (right) pair.}}
    \label{fig:FET_pair}
\end{figure}
\subsubsection{Face Examiner Test}
Baseline performance on facial identification was measured with the FET \cite{phillips2018face}, which contains 20 pairs of images (12 same-identity and 8 different-identity) taken in indoor and outdoor environments. Image pairs consist of White (18 image pairs) and East Asian (2 image pairs) faces (see \cite{phillips2018face} for complete details of the FET). Figure \ref{fig:FET_pair} depicts two example face image pairs from this test.

\section{Results}\label{sec:results}
\subsection{Cross-Race Test}
Figure \ref{fig:short} shows performance on the Cross-Race Test for humans and machines.  Machine performance in all conditions was perfect, as indicated by the red line on the figure at AUC = 1.0. 

For humans, accuracy was above chance for all conditions. Notably, performance varied considerably for individual participants in all conditions, as indicated by the spread
of points within each violin plot in Figure \ref{fig:short}. By design, the performance proved roughly comparable for participants of both races on face-image pairs of Black and White individuals median AUCs: Black participants (Black face-image pairs = 0.67, White face-image pairs = 0.65); White participants (Black face-image pairs = 0.64; White face-image pairs = 0.60). 

These observations were confirmed formally by submitting the human AUC data to a two-way mixed model analysis of variance (ANOVA) with participant race (Black, White) as a between-subjects variable and stimulus race (Black, White) as a within-subjects variable. An ANOVA is conducted to analyze the statistical difference between groups of subjects. A $p$-value that is less than $0.05$ indicates that differences between groups are statistically significant. No factors proved significant statistically ($p > 0.05$) in all cases. Therefore, Black and White face-image pairs were  equal difficulties for Black and White participants.



\subsection{Face Examiner Test Results}\label{NIST-UTD_FET_results}

\begin{figure}
    \centering
    \begin{tabular}{c}
            \includegraphics[width=0.50\textwidth]{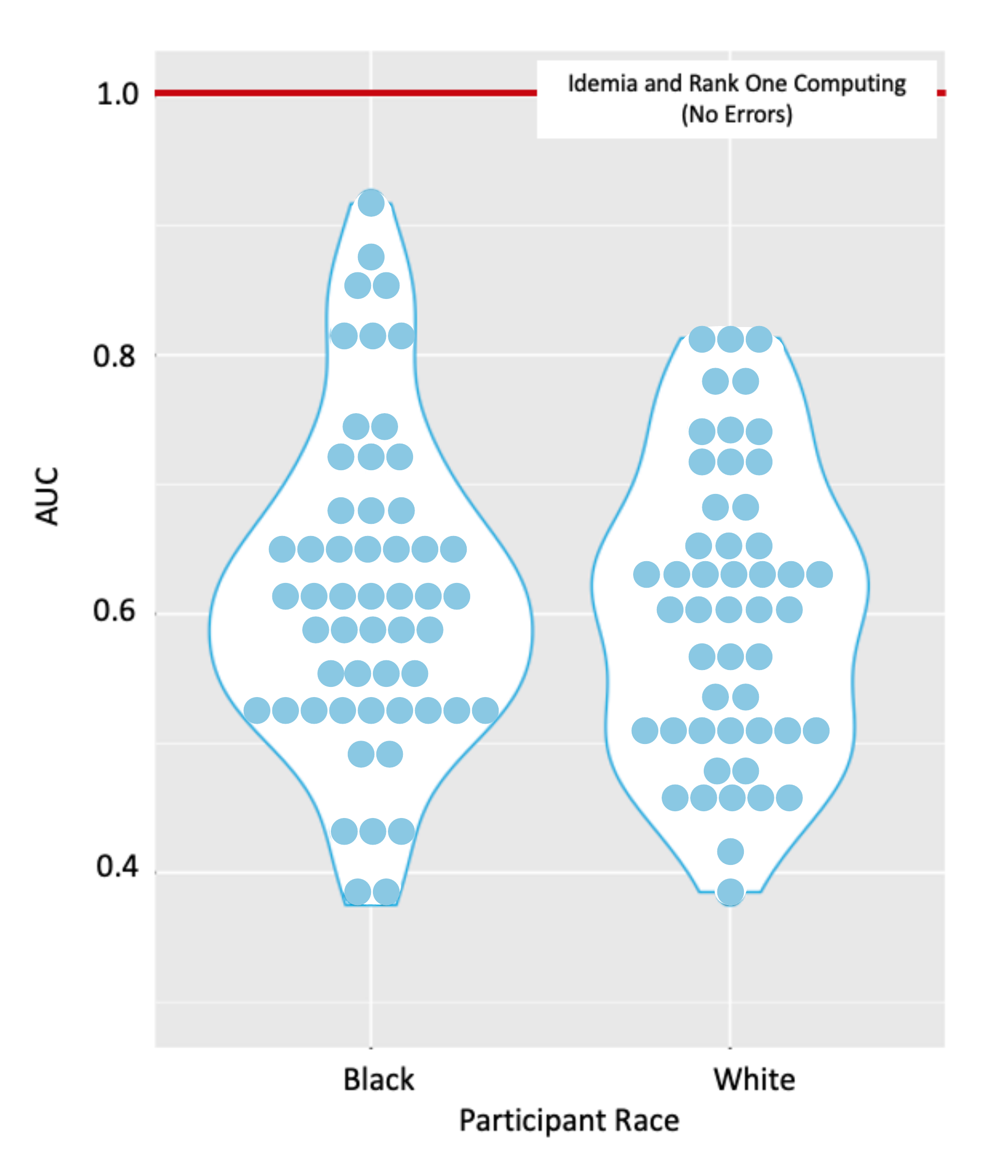} 
    \end{tabular}
    \caption{AUC values of untrained participants on the FET. The $x-axis$ depicts the race of the participants and the $y-axis$ shows the AUC performance on the FET test. The median AUC value for both groups of human subjects was $0.604$. The red line indicates the performance of two face verification systems from the FRVT evaluation, Idemia and Rank One Computing Submissions, which both had AUC values of $1.0$.
    }
    \label{fig:bb_auc}
\end{figure}

Figure \ref{fig:bb_auc} shows human and machine performance on the FET. For machines, again, performance was perfect, as indicated by the red line at the top of the figure.

For humans, performance on the FET (Median AUC = 0.604)  in this study
was far from perfect and  roughly comparable to untrained participants tested in Phillips et al. \cite{phillips2018face} on the FET (AUC = 0.67). Like the Cross-Race test, performance did not differ for Black and White participants
($p =$ 0.51). 

\section{Conclusions}

Between 1993 and 2022, the error rate for face recognition systems has halved every two years for mugshot style images \cite{Phillips2011IdentityScience,Phillips:FedID2022}.
These performance gains are readily apparent in ongoing comparisons between human and machine performance.

The literature has shown that even with the substantial performance improvements seen, including in the deep learning era, race bias remains a concern \cite{cavazos2020accuracy,grother2019face}.
It is important to understand that the present 
results do not demonstrate that state-of-the-art face recognition 
is now racially equitable. Instead, we found a specific condition under which two algorithms are superior to humans when comparing Black and White face pairs.

Turning to humans, we tested untrained participants as a proxy for the general population.
Previous studies have shown that the face recognition ability of students equals that of police officers and passport officers \cite{Burton1999FaceRI, White2014PassportOE}. This suggests there exists identity-checking applications where state-of-the-art face verification systems surpass humans. These include applications that check the identity of multiple races. 

In further constraining the conclusions of this work,
the face verification test we developed here used only  frontal images.
Face recognition systems  are more  accurate with  frontal images than with in-the-wild images that vary substantially in quality, pose (yaw, pitch), illumination, and expression.
It remains an open question whether the present results  generalize to more challenging face recognition applications with in-the-wild images.

In summary, this study presents the first human-machine comparison for cross-race face identification in the era of deep learning. Results indicate that under the conditions we tested, face recognition systems are highly accurate 
on a task that humans found highly challenging. The cross-race test we developed here can be useful for evaluating human experts, including trained face forensic professionals (examiners, reviewers) and super-recognizers to benchmark them against current face recognition systems. 

\section*{Disclaimer} Certain equipment, instruments, software, or materials are identified in this paper in order to specify the experimental procedure adequately.  Such identification is not intended to imply recommendation or endorsement of any product or service by NIST, nor is it intended to imply that the materials or equipment identified are necessarily the best available for the purpose.

\section*{Acknowledgment}

The authors would like to thank Kayee Hanaoka for running the FRVT ongoing systems on our images, and the NIST Forensic Science Research Program for supporting this project.  

Research at the University of Texas at Dallas was funded by The National Institute of Standards and Technology, the National Eye Institute Grant R01EY029692-04 to AOT, Grant 70NANB21H109 \& 70NANB22H150 to A.OT. Corresponding author (GJ) supported
in part by National Institute Standards and Technology Grant 70NANB21H109 to AOT.

{\small
\bibliographystyle{ieee}
\bibliography{references}
}

\end{document}